\documentclass[runningheads]{llncs}
\usepackage{afterpage}
\usepackage{algorithm,algorithmic}
\usepackage{array} 
\usepackage{amsmath,amssymb,amsfonts}
\usepackage{booktabs} 
\usepackage{color}
\usepackage{graphbox}
\usepackage{graphicx}

\usepackage{mathtools}
\usepackage{multirow}
\usepackage{tabularx}
\usepackage{textcomp}
\usepackage{tikz}
\usepackage{pifont} 
\usepackage{colortbl}
\usepackage[normalem]{ulem}
\useunder{\uline}{\ul}{}
\usepackage{hyperref}
\usepackage{cleveref} 
\usepackage{atbegshi} 

\usepackage{bibentry}
\usepackage{longtable}
\usepackage{csquotes}

\usepackage{flushend}
\usepackage{cleveref}
\usepackage{subcaption}
\usepackage{graphicx}
\usepackage{orcidlink}
\usepackage{changepage}

\usepackage{pgfplots}
\usepackage{src/tkz-kiviat} 
\usetikzlibrary{arrows}

\usepackage[nomessages]{fp}
\usepackage{censor}
\usepackage[font=small,skip=6pt]{caption}
\usepackage{lmodern}
\usepackage{relsize}

\newcounter{ch}
\usepackage{array,ragged2e}
\newcolumntype{P}[1]{>ŵ{\RaggedRight\arraybackslash}p{#1}}
\usepackage[T1]{fontenc}
\usepackage{todonotes}

\AtBeginShipout{\ifnum\value{page}=13\pagecolor{red!20}\fi}
\begin{document}
\title{Multi-dimensional Autoscaling of Processing Services: A Comparison of Agent-based Methods}
%
\titlerunning{Multi-dimensional Autoscaling of Processing Services: Agent-based Methods}
%


\author{Boris Sedlak\inst{1,*}\orcidlink{0009-0001-2365-8265} \and
Alireza Furutanpey\inst{1}\orcidlink{0000-0001-5621-7899} \and
Zihang Wang\inst{1}\orcidlink{0009-0009-8194-4698} \and
Víctor Casamayor Pujol\inst{2}\orcidlink{0000-0003-2830-8368} \and
Schahram Dustdar\inst{1,2}\orcidlink{0000-0001-6872-8821}}
\authorrunning{Sedlak et al.}
%
\institute{Distributed Systems Group, TU Wien, Vienna, Austria.
\and
DISL, ICREA | Universitat Pompeu Fabra, Barcelona, Spain\\
* Corresponding author contact: \email{b.sedlak@dsg.tuwien.ac.at} 
}
\maketitle 
\begin{abstract}

Edge computing breaks with traditional autoscaling due to strict resource constraints, thus, motivating more flexible scaling behaviors using multiple elasticity dimensions. This work introduces an agent-based autoscaling framework that dynamically adjusts both hardware resources and internal service configurations to maximize requirements fulfillment in constrained environments. We compare four types of scaling agents: Active Inference, Deep Q Network, Analysis of Structural Knowledge, and Deep Active Inference, using two real-world processing services running in parallel: YOLOv8 for visual recognition and OpenCV for QR code detection. Results show all agents achieve acceptable SLO performance with varying convergence patterns. While the Deep Q Network benefits from pre-training, the structural analysis converges quickly, and the deep active inference agent combines theoretical foundations with practical scalability advantages. Our findings provide evidence for the viability of multi-dimensional agent-based autoscaling for edge environments and encourage future work in this research direction. 
\keywords{
Internet of Things 
\and Stream Processing 
\and Active Inference 
\and Autoscaling 
\and Markov Decision Processes 
\and Reinforcement Learning }
\end{abstract}

\section{Introduction}\label{sec:introduction}
\begin{figure}[t]
    \centering
    \begin{subfigure}[b]{0.48\linewidth}
        \includegraphics[width=\linewidth, trim=200 111 0 0, clip]{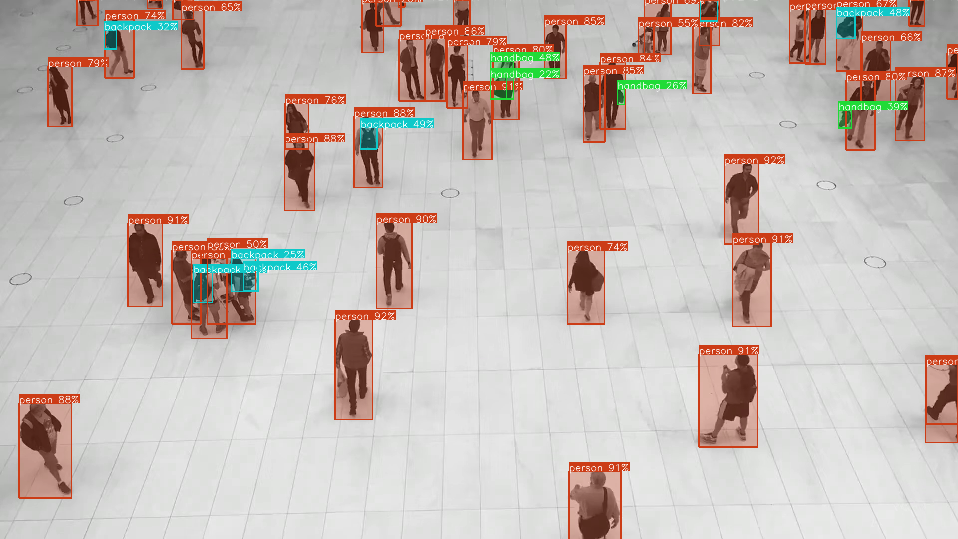}
        \caption{CV (Yolov8)}
        \label{fig:cv-demo}
    \end{subfigure}
    \hfill
    \begin{subfigure}[b]{0.48\linewidth}
        \includegraphics[width=\linewidth]{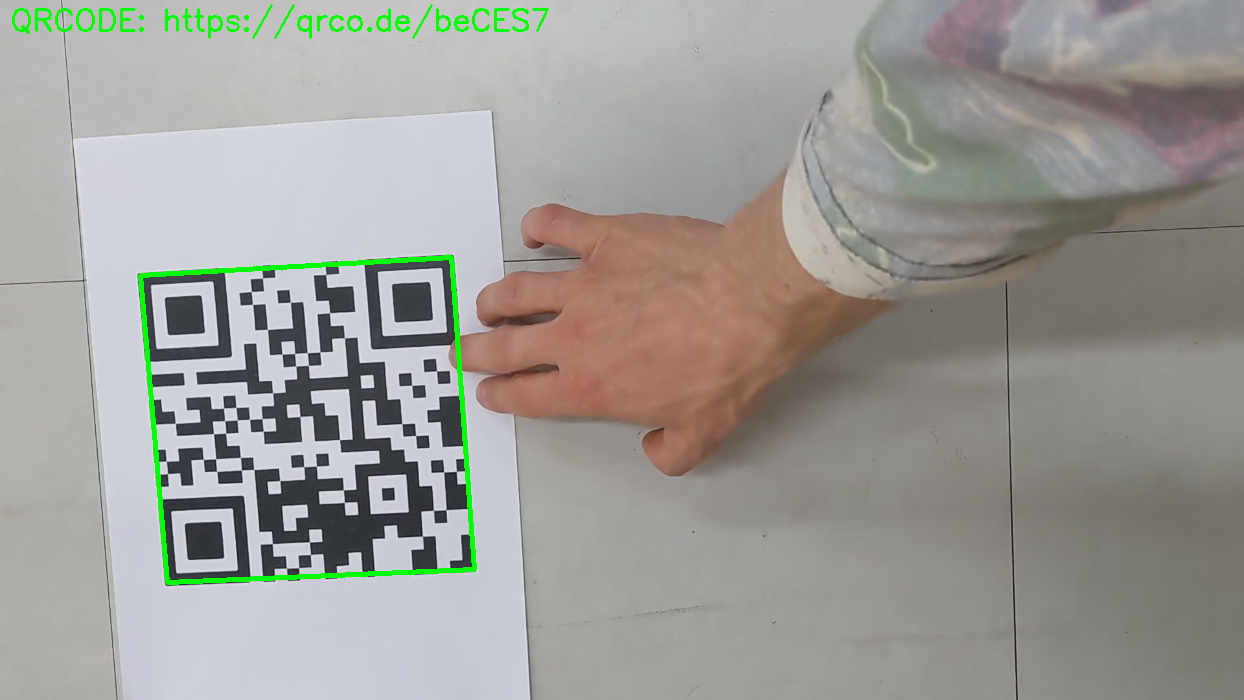}
        \caption{QR (OpenCV)}
        \label{fig:qr-demo}
    \end{subfigure}
    \caption{Demo output of the results produced by the two processing services}
    \label{fig:demos}
    \vspace{-5pt}
\end{figure}

The rise of Edge Computing and the Computing Continuum (CC) addresses the limitations of traditional Cloud infrastructures~\cite{dustdar_distributed_2023}. By bringing computation closer to users and data sources (e.g., IoT devices) these paradigms significantly reduce network latency, critical for applications that demand near real-time responses, such as autonomous driving, e-health, and virtual reality. A common use case, as depicted in Figure~\ref{fig:demos}, could be to detect entities in a video stream (e.g., humans) or tracking objects that have a QR code attached. By running these inference services locally, the overall network congestion is also mitigated by minimizing long-distance data transfers. 

However, Edge and CC environments introduce new challenges \cite{casamayor-pujol_fundamental_2023}: They rely on resource-constrained computing hardware, and thus break with traditional Cloud-based autoscaling. In Cloud systems, autoscaling mechanisms elastically respond to increased user demand by allocating more resources to a service or replicating it. This is infeasible at the Edge or in the CC, where computing resources are strictly limited \cite{manaouil_kubernetes_2020}. Especially when resources are scarce, applications require a more flexible scaling behavior that uses a wider range of adaptations -- hence, operating in multiple \textit{elasticity dimensions}~\cite{dustdar_principles_2011}. 
On the one hand, this protects the service execution and promises higher requirements fulfillment -- captured through a set of Service Level Objectives (SLOs). On the other hand, this increases the complexity for choosing optimal scaling actions. What is needed, hence, are lightweight multi-dimensional scaling mechanisms that optimize the service execution without obstructing existing workloads.

To fill this gap, we propose an agent-based autoscaling approach tailored for Edge and CC systems, which adjusts processing services in multiple elasticity dimensions. Our approach employs decentralized local agents that (1) observe the service execution and their SLO fulfillment without centralized control; thus, we can monitor the resource allocation per service or the application throughput. If SLOs are violated, the agents attempt to restore the desired state by (2) adjusting the service execution; the exact scaling policy is learned by the agent according to environmental feedback. Notably, our approach allows scaling policies tailored to the individual services, where one service could, for example, scale down its machine learning (ML) model, while another service claim the remaining resources. This allows building composite and customizable scaling policies, which go further than existing approaches~\cite{sedlak_towards_2025_short}.
%

To show the viability of our approach, we implement four different versions of our general agent and compare their performance in a processing environment, where the agent needs to dynamically scale two processing services on an Edge device. In particular, we compare an Active Inference agent (AIF), a Deep Active Inference agent (DACI), a Deep Q-Network agent (DQN), and an agent using a numerical solver -- called Analysis of Structural Knowledge (ASK). During our experiments, a scaling agent manages two physically executed services: one for video stream inference (Figure~\ref{fig:cv-demo}) using the well-known Yolov8 model~\cite{varghese_yolov8_2024}, and another for QR code reading (Figure~\ref{fig:qr-demo}), implemented with OpenCV2~\cite{opencv_opencv_2024}.

On the short term, our work provides well-needed baselines for comparing the performance of different agent-based approaches -- particularly needed for emerging solutions. On the long term, our evaluation environment is extensible and allows incorporating other agents.
%
We summarize our contributions as:
%
\begin{enumerate}
    \item Introducing an agent-based autoscaling approach within a \textit{multi-dimensional elasticity space} that dynamically maximize SLO fulfillment in IoT and Edge environments by adjusting hardware and service configurations
    \item Evaluating on \textit{real-world applications} of four distinct scaling agent architectures (Active Inference, Deep Active Inference, Deep Q-Network, and Analysis of Structural Knowledge) for real-time service orchestration.
    \item Providing a benchmarking environment for future autonomous research and demonstrating viability through agent-based resource allocation for parallel processing services (YOLOv8 and OpenCV) on constrained hardware.
\end{enumerate}
%
%
%
%
%
%
%
%
%
%
%

\section{Preliminaries}\label{sec:preliminaries}

In the following, we provide a formal description of the problem, as well as how researchers have addressed it so far with agent-based methods -- including AIF.

\subsection{Problem Definition}
\label{subsec:problem-definition}


As depicted in Figure~\ref{fig:simplified-problem}, multiple services are executed within one Edge device -- sharing the device's processing resources between them. The execution of the individual services is affected by the amount of resources allocated (e.g., CPU, or RAM), and the configuration of the service-internal parameters (e.g., input quality, or model size). Considering that both sets of parameters can be \textit{elastically} adjusted during runtime~\cite{dustdar_principles_2011}, we summarize them as elastic configurations.

The allocated resources and the service configuration influence the degree to which the service outcome satisfies the client requirements (i.e., the SLOs). However, and this is the core of the problem, it is not a priori known what will be the resulting SLO fulfillment for a specific configuration.
Hence, the problem boils down to adjusting the elastic configurations in such a way, that the SLO fulfillment is maximized.
While the number of service configurations and resource allocations is limited, it is not easily possible to brute-force the problem by exhaustively searching the solution space. The reason is that actions taken on the environment require a considerable amount of time to show effect. For instance, orchestration tools like Docker and Kubernetes usually consider a cooldown period of several minutes after taking an action. 

\subsubsection{Formal Definition}

More formally, the problem domain and the physical processing environment is defined as follows:

\begin{figure}[t]
    \centering
    \includegraphics[width=1\linewidth]{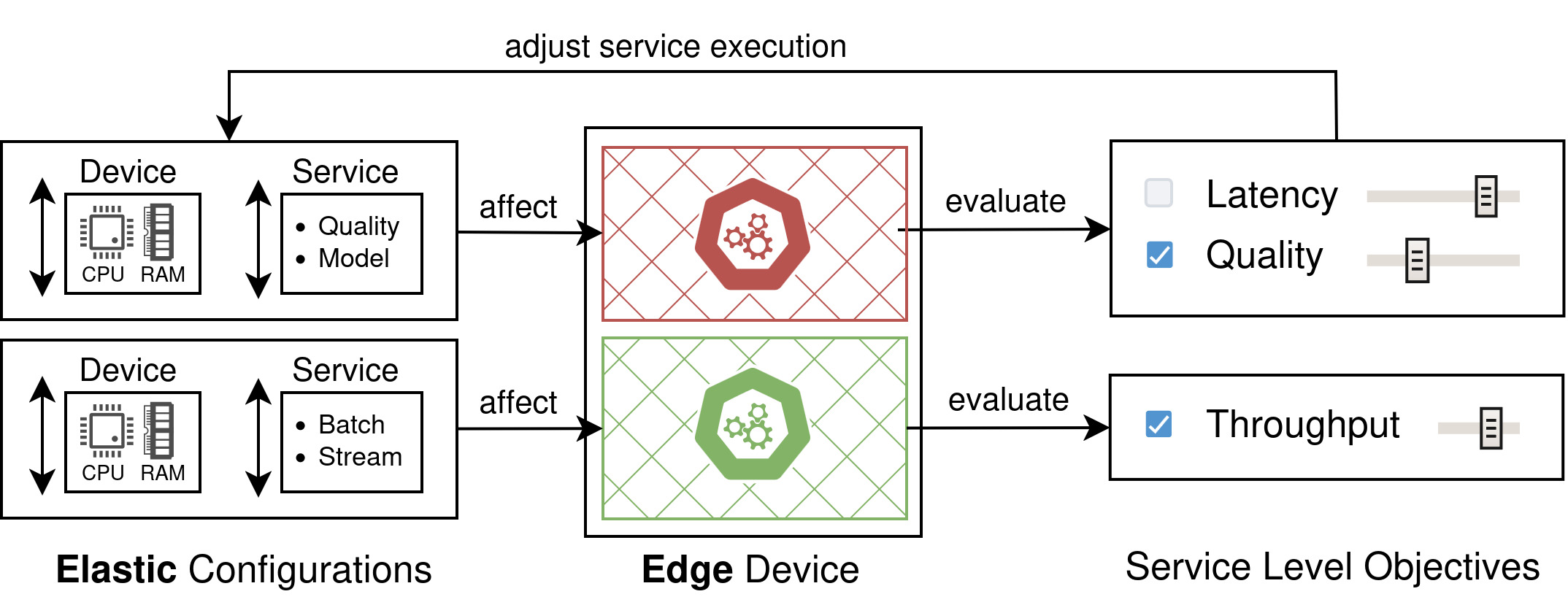}
    \caption{Services compete for limited device resources; according to SLO fulfillment, the service execution is adjusted by elastically changing device and service configurations}
    \label{fig:simplified-problem}
    \vspace{-5pt}
\end{figure}

\paragraph{Processing Hardware}

An Edge device $d$ is defined by its hardware constrains $H$, e.g., the physical number of CPU cores $c_{phy}$ and RAM capacity $r_{phy}$, and the set of processing services $S$ that is executed there; hence $d = \langle H,S \rangle$.

\paragraph{Processing Services}

Each processing service ($s \in S$) is characterized through a service type $t$, a set of SLOs $Q$, and a service-internal configuration $K$; hence $s = \langle t, Q, K, c_s \rangle$, where a service (s) has a number of cores ($c_s$) allocated to it, for which $c_s \leq c_{phy}$ must hold. Each SLOs $q\in Q$ with $q = \langle v,t \rangle$ tracks a variable $v$ and reflects whether its current assignment reaches a target ($t$).


\paragraph{Service Monitoring}

To track the system state and the SLO fulfillment, the device $d$ and its services $s \in S$ continuously monitor the execution at different levels. This provides a set of software- or application-related metrics ($M_s$), and a set of device-related metrics ($M_d$) that capture the hardware state.

\paragraph{SLO Fulfillment}

Using these metrics, we calculate the continuous SLO fulfillment ($\phi$) for a metrics ($m \in M$) and an SLO $q$ as shown in Eq.~\eqref{eq:slo-f}.
\begin{equation}
\phi(q, m) = 
\begin{cases}
\frac{m}{t_q} & \text{if } m \leq t_q \\
1.0 & \text{if } m > t_q
\end{cases}
\label{eq:slo-f}
\end{equation}
%
This means, that SLOs cannot be overfulfilled; for instance, if the target is keeping the service throughput ($tp$) $\geq 30$, both assignments for $m_\textit{tp} = 40$ and $m_\textit{tp} = 100$ achieve the maximum SLO fulfillment of $\phi = 1.0$.



\subsection{Related Work}
\label{subsec:related-work}
In the following, we identify competing approaches on autoscaling that do not use AIF, and others that use AIF agents.

Most notably, competing approaches and traditional autoscaling methods~\cite{wang_autothrottle_2024,zhao_tiny_2022_short} struggle in environments, where multi-dimensional adaptation is crucial for maintaining service quality. In particular, RL-based systems still find no productive use by large providers despite over a decade of research. We argue that AIF-based agents offer a promising alternative by enabling adaptive, efficient, and decentralized control suited to the dynamic and uncertain nature of complex multi-tier distributed systems. Hence, our focus is on further establishing AIF for service adaptation on resource-restricted devices.
%
%

Among existing works, Sedlak et al.~\cite{sedlak_equilibrium_2024} explore how AIF can be used to optimize SLO fulfillment across various use cases. Danilenka et al.~\cite{danilenka_adaptive_2024_short} demonstrate how AIF agents can adapt to dynamic and heterogenous environments.
%
Pujol et al.~\cite{pujol_distributed_2025} adopt a representation using Partially Observable Markov Decision Processes (POMDPs) within a multi-agent system.
%
Lapkovskis et al.~\cite{lapkovskis_benchmarking_2025} presented a comparison of AIF agents with other methods for a CC application that complements our work with a different set of algorithms. 
Vyas et al.~\cite{vyas_towards_2025} provide an adaptation mechanism for active sensing on Edge devices, which dynamically adjust perception by changing the camera orientation.

\section{Methodology}\label{sec:methodology}
In this section, we first present the general design of a scaling agent that maximizes the SLO fulfillment for a set of services under limited processing resources.
Afterward, we show four approaches of how this agent can be implemented. Namely, we present a traditional RL agent using a Deep Q-Network (DQN), an Bayesian agent using AIF, a Deep Active Inference (DACI) agent, and an algebraic agent based on Analysis of Structural Knowledge (ASK).
\subsection{Processing Environment}
To evaluate our scaling agents for a real-world problem, we need a processing environment in which to operate. For this, we first introduce the structure of the two processing services and how they are monitored during runtime.
%
Lastly, we introduce our training environment for pretraining some of the agents.
\subsubsection{Processing Services} 

In the following, we introduce the two processing services that will be executed in parallel on an Edge device. By limiting ourselves to two services, we decrease the complexity of the optimization problem, while retaining the central issue of resource limitations.


\paragraph{Computer Vision (CV)}

The CV service processes a continuous stream of video frames to perform object detection on them. For this, it uses Yolov8~\cite{varghese_yolov8_2024}, a DNN model developed by Ultralytics; Figure~\ref{fig:cv-demo} showcases its output.

The service state for CV is composed by four features, as displayed in Table~\ref{tab:CV-features}: \textit{quality} determines the ingested video resolution, \textit{model size} describes the Yolo model (e.g., v8n or v8m), and \textit{cores} determines the maximum resources allocated; \textit{throughput} describes the service output in terms of frames per second.

The action state for CV is a subset of these variables -- as part of the service configuration ($K$), it is possible to adjust \textit{quality} and \textit{model size}, while \textit{cores} ($c_s$) is another property of the service. The exception is \textit{throughout}, which cannot be directly set, but is statistically dependent on the three other features.
\begin{table}[h]
\setlength{\tabcolsep}{6pt}
    \vspace{-8pt}
    \centering
    \begin{tabular}{r|cc|cc|c}
        variable & type & range & actionable & dependent & SLO target\\
        \hline
        \textit{quality} & int & [128, 320] & $\checkmark$ (step$=32$) & --- & $\geq 288$\\
        \textit{model size} & int & [1, 5] & $\checkmark$ (step=1) & --- & $\geq 3$\\
        \textit{cores} & float & (0, 8)& $\checkmark$ (no step) & --- & ---\\
        \textit{throughput} & int & [0, 100] & --- & all others & $\geq 5$\\
    \end{tabular}
    \vspace{4.5pt}
    \caption{Variables of the CV service used for sensing and acting on the environment}
    \label{tab:CV-features}
    \vspace{-20pt}
\end{table}

\paragraph{QR Code Reader (QR)}

The QR service scans a continuous video stream to detect QR codes within the individual video frames. For this, it uses the Python wrapper of OpenCV~\cite{olli-pekka_heinisuo_opencv-python_2021}; Figure~\ref{fig:qr-demo} showcases the service output.

The service state for QR is composed by three features, as displayed in Table~\ref{tab:QR-features}. The three features are defined analog to the CV service; the exception is that QR does not use a specific model size but a fixed algorithm.
%
%
\begin{table}[h]
\setlength{\tabcolsep}{6pt}
    \vspace{-8pt}
    \centering
    \begin{tabular}{r|cc|cc|c}
        variable & type & range & actionable & dependent & SLO target\\
        \hline
        \textit{quality} & int & [300, 1000] & $\checkmark$ (step$=100$) & --- & $\geq 900$\\
        \textit{cores} & float & (0, 8)& $\checkmark$ (no step) & --- & ---\\
        \textit{throughput} & int & [0, 100] & --- & all others & $\geq 60$\\
    \end{tabular}
    \vspace{4.5pt}
    \caption{Variables of the QR service used for sensing and acting on the environment}
    \label{tab:QR-features}
    \vspace{-20pt}
\end{table}

To quantify the requirements and preferences on how the services should operate, both Table~\ref{tab:CV-features} \& \ref{tab:QR-features} contain the precise SLO targets. Despite the fact that both services operate on a video stream, they differ greatly in their input shape and the expected \textit{throughput}. Notably, QR has a high expected throughput of 60 frames per second, while the resource-heavy CV has a target of 5.

\subsubsection{Service Monitoring}
\label{subsubsec:monitoring}


The two processing services operate in batches of 1 second: at the beginning of each iteration, 100 frames are ingested to the service; when the processing timeframe (i.e., 1000 ms) is exceeded, the service counts the number of processed frames -- the \textit{throughput}. This information, together with the service properties, is then collected in a time-series DB. 

Later, when a scaling agent wants to resolve the service states, it can query this information through the time-series DB. A key advantage of this approach, is that it allows computing sliding-windows over monitored variables; thus, it stabilizes an agent's perception against temporary perturbations, like momentary drops in throughput. Stable states are most relevant for evaluating SLOs and avoiding overhasty scaling decisions, e.g., the Kubernetes HPA\footnote{https://kubernetes.io/docs/tasks/run-application/horizontal-pod-autoscale/} considers per default a time window of 30min; in our case, we will stick to 5 seconds. 

\subsubsection{Training Environment}
\label{subsubsec:training-env}

Changes to the processing environment need time to show effect. This, however, is conflictive with contemporary ML training, which often requires thousands of iteration to converge to a policy. Hence, sample-efficient methods, like model-based algorithms (also evaluated in this paper) move into focus. However, to not exclude traditional RL algorithms, like DQN, we supply a training environment based on the Gymnasium framework\footnote{https://gymnasium.farama.org/api/env/}.

To estimate the state transition probabilities for state-action pairs, we use a part of the monitored metrics to train a Linear Gaussian Bayesian Network (LGBN); this can be seen as a simplified replication of the real processing environment. For an assignment of free (i.e., actionable) variables, the LGBN samples the remaining variables -- in this case \textit{throughput}. Training on the LGBN environment, scaling agents can infer expected state and reward of actions. 

\subsection{General Agent Design}
\label{subsec:general-agent}

To optimize SLO fulfillment, we supervise the service execution through a dedicated scaling agent, executed locally on the processing device.
%
Generally, our agent follows a simple two-step scheme in which it first resolve the states of services and hardware, and then acts on the processing environment. To react to dynamic runtime behavior, our scaling agents iterate in cycles of 5 seconds -- thus adhering to the sliding window for service monitoring.


\subsubsection{Perception}

At the beginning of each iteration, the agent queries the service states through the time-series DB (cfr. Section~\ref{subsubsec:monitoring}).
%
%
Next, it iterates through the services ($s \in S$) and evaluates their SLO fulfillment as in Eq.~\eqref{eq:slo-f}. This indicates the degree to which requirements are currently fulfilled.

To determine the amount of unclaimed resources, the agent then resolves the number of cores ($c_s$) currently assigned to the individual services\footnotemark and the number of free cores ($c_{free}$) on the device $d$ as in Eq.~\eqref{eq:free}. 
\begin{equation}
    \label{eq:free}
    c_{free} = c_{phy} - \sum_{s \in S} c_s .
\end{equation}
\vspace{-15pt}
%
%
\footnotetext{Internally, the services are executed in a containerized environment -- Docker; the claimed cores per service can be queried by our agent running on the same machine.}

\subsubsection{Action}

After resolving the environmental state, the agent adjusts the processing environment according to its preferences by changing actionable service variables. However, the exact policy depends on the implementation of the scaling agent and will be explained in the next section. 
For example, some agents will try solving this by training an accurate world model, e.g., the AIF agent, while the DQN will use simplified state-action networks.
Most importantly, this also determines how the different scaling agents explore the solution space.

\subsection{Scaling Agent Implementations}
\label{subsec:agent-impl}

In the following, we describe four different implementations of the general scaling agent, using different agent-based approaches. The code for the agents, as well as for the processing services, can be found in the following repository\footnotemark.

\footnotetext{\href{https://github.com/borissedlak/elastic-workbench/tree/iwai-camera-ready}{Repository} with implementations for scaling agents and processing services}



\subsubsection{Active Inference (AIF)}

The AIF agent is defined following the same idea as in~\cite{pujol_distributed_2025}, which leverages \texttt{pymdp}~\cite{heins_pymdp_2022}. Hence, we model the agent as a Markov Decision Process and specifying the transition model using Dynamic Bayesian Networks and Conditional Probability Tables (CPTs). In particular, the influence of actions over certain state factors, such as \textit{throughput}, whose transition dynamics are unknown, is initially defined as a uniform probability distribution within the CPTs\footnotemark. These uniform priors allow the agent to learn the true dynamics through experience and improve SLO fulfillment.

\footnotetext{The interested reader can find the matrix definition in the following \href{https://github.com/borissedlak/elastic-workbench/blob/iwai-camera-ready/iwai/pymdp_agent.py}{directory}.}

The AIF agent simultaneously controls both services, which increases its state and action spaces. To maintain practical computational times, the agent can choose only one action per service: modifying data \textit{quality}, \textit{model size}, or number of \textit{cores}. This reduces computational demands but also restricts the agent’s capacity for more dynamic and fine-grained interventions. Additionally, compared to the other agents in this article, the state space factors for this AIF agent are discretized into coarser bins. While this lowers computational cost, it results in less precise action selection and coarser estimates of SLO fulfillment. The final state space of the agent consists of 7 factors with 3,457,440 state combinations, and an action space of 35 different combinations.


Using this setup, the pymdp implementation (using v0.0.7.1) remained prohibitively slow, requiring approximately 20 seconds per iteration for policy inference and observation model construction. We address these bottlenecks through two algorithmic optimizations: (1) vectorized policy evaluation eliminates nested loops in expected free energy calculations, and (2) sparse matrix operations reduce overhead in belief updating. These improvements achieved 20-30x speedup in inference routines, leading to the performance describing in Figure~\ref{fig:loop_time}.

%
\subsubsection{Deep Active Inference (DACI)}
While traditional AIF implementations using graphical models provide excellent interpretability, they are challenging to scale, limiting their practicality in real-world applications with high-dimensional input spaces. DACI addresses scalability in the CC by exploiting the abundant availability of hardware acceleration (e.g., TPUs, NPUs) for Artificial Neural Networks (ANNs).
This work presents preliminary results on Deep Active Inference for the CC based on the work by Fountas et al~\cite{mcdaci}. The basic idea is to learn non-linear transforms with ANNs to map the high-dimensional input space into a compressed latent representation of the environment. Then, the agent may operate efficiently in complex, multi-dimensional elasticity spaces that would be intractable for discrete graphical models. Analogous to any active inference agent, it 1) samples the environment and calibrates its internal generative model to explain sensory observations and 2) performs actions to reduce the uncertainty about the environment.
We slightly simplify and adjust the original objective in \cite{mcdaci} to compute the variational free energy for each time-step \textit{t} as:
\begin{align*}
\mathcal{F}^{\prime}_{t} =\ & \mathbb{E}_{Q_{\phi_s}(s_t)}\left[\log P_{\theta_o}(o_t|s_t)\right] \\
&+ D_{KL}\left[Q_{\phi_s}(s_t)\,\|\,P_{\theta_s}(s_t|s_{t-1}, a_{t-1})\right] \\
&+ \lambda_{\text{eq}} \cdot \text{Var}[\text{fulfillment}_{\text{services}}] \\
&+ \lambda_{\text{util}} \cdot \left(1 - \frac{\text{ucores}}{\text{tcores}}\right)^2
\end{align*}
We have omitted the KL divergence involving the habitual network and included regularization terms to reduce the variance between service fulfillments and encourage resource utilization. Our intuition is to regularize the term $-\mathbb{E}[\log P(o_\tau|\pi)]$ that represents preferences about future observations by encoding the specification for balanced services and resource efficiency to have a higher prior probability.
Lastly, we encode the objective to prioritize actions that reduce uncertainty about the environment by minimizing the EFE with an expression that is equivalent to that in the original work~\cite{mcdaci}.
\subsubsection{Deep Q Networks (DQN)}
 The DQN~\cite{dqn} agent approximates Q-values for state-action pairs in the multi-dimensional autoscaling problem. We implement separate DQN models for each processing service (CV and QR), allowing service-specific policy learning while maintaining coordinated resource allocation when competing for limited resources. The models are trained jointly within the shared processing environment to benefit from the pre-training in the simulated LGBN environment, resulting in the DQN agent achieving stable performance from the initial deployment phase.

\subsubsection{Analysis of Structural Knowledge (ASK)} 
This agent solves the orchestration problem through numerical optimization, using SLSQP. For this, it consider the variables and their parameter boundaries from Table~\ref{tab:CV-features} \& \ref{tab:QR-features}, the objective function $\phi$ for calculating the SLO fulfillment of a parameter assignment, and the SLOs ($Q$). Given that, it misses a continuous function that allows descending to the optimal solution. In a nutshell: how to estimate \textit{throughput}, given the assignments of all bounded parameters -- including continuous-scale \textit{cores}?

We address this, similarly to the LGBN training environment, through a regression model that captures the dependencies between service variables ($K$) and the allocated cores ($c_k$). As the agent captures more metrics, the accuracy of the regression and the inferred parameter assignment improves. To create an accurate model, the ASK agent needs an initial exploration time (i.e., 20 iterations in the experiments), during which the parameters are assigned randomly. After this time, the ASK is ready for the numerical optimization.

For our two service $s_1$ and $s_2$ and all their variables $x \in \{ K_s \cup c_s \}$, the ASK agent infers assignments that maximize the SLO fulfillment as in Eq.~\eqref{eq:global-slo-optimization}. 

\begin{equation}
\max_{\mathbf{x}} \sum_{s \in S} \phi_s(x) \quad \text{subject to} \quad \sum_{s \in S}\text{c}_s \leq c_{\text{phy}}, \quad x_i \in [x_i^{\min}, x_i^{\max}]
\label{eq:global-slo-optimization}
\end{equation}


This presents a couple of advantages: First, the continuous action space allows agents to infer fine-grained elasticity strategies; in particular \textit{cores} can be split precisely between the devices. Also, it is possible to infer assignments for all services and all their elasticity parameters in one operation.


\section{Evaluation}\label{sec:evaluation}

\subsection{Experimental Design}

To analyze the performance of the different scaling agents we evaluate them, one after the other, in the processing environment (i.e., the two processing services executed in parallel). 
In each experiment we capture the following performance metrics: (1) SLO fulfillment, as the average over the two services, and (2) complexity of inference, as the duration for running one inference cycle. 

Each experiment contains 50 iterations within the processing environment, i.e., 50 times that the agent interacts with the environment by sensing its state and acting on it. As described in Section~\ref{subsec:general-agent}, the agents operate with a frequency of 5 seconds -- every 5 seconds they orchestrate the services according to the inferred policy. Thus, individual experiments take $5 \times 50 = 250$ seconds. To stabilize the empirical results, we repeat each experiment 10 times.

During the experiments, agents must achieve the SLO targets specified in Table~\ref{tab:CV-features} \& \ref{tab:QR-features}. To fulfill these SLOs, an agent needs to optimally adjust the services and split the resources -- in this case 8 physical cores (i.e., $c_{phy} = 8$).

\subsection{Experimental Results}



\begin{figure}[t]
    \centering
    \begin{subfigure}[b]{0.469\linewidth}
        \includegraphics[width=\linewidth]{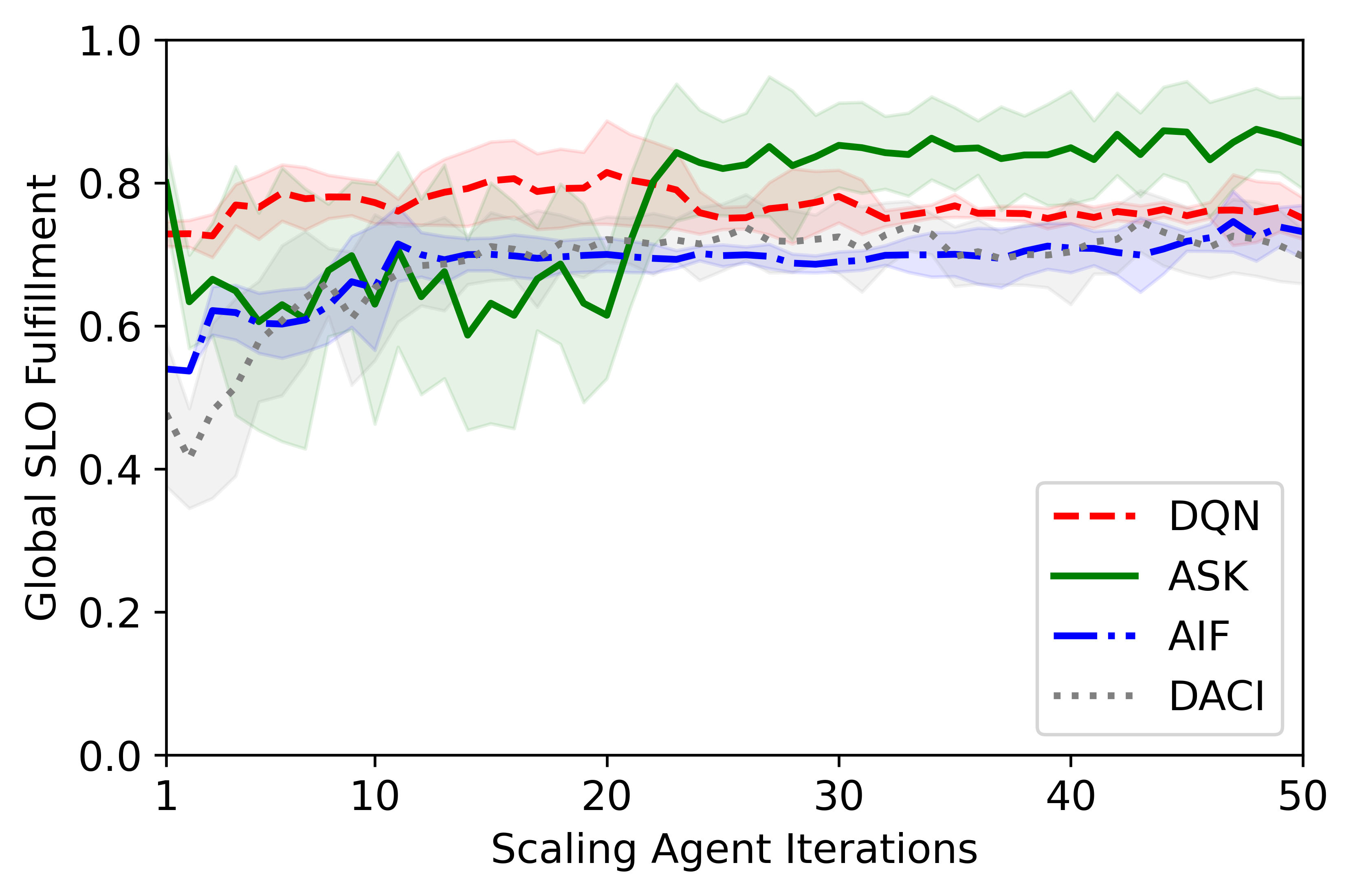}
        \caption{SLO fulfillment of different agents}
        \label{fig:slo_fulfillment}
    \end{subfigure}
    \hfill
    \begin{subfigure}[b]{0.491\linewidth}
        \includegraphics[width=\linewidth]{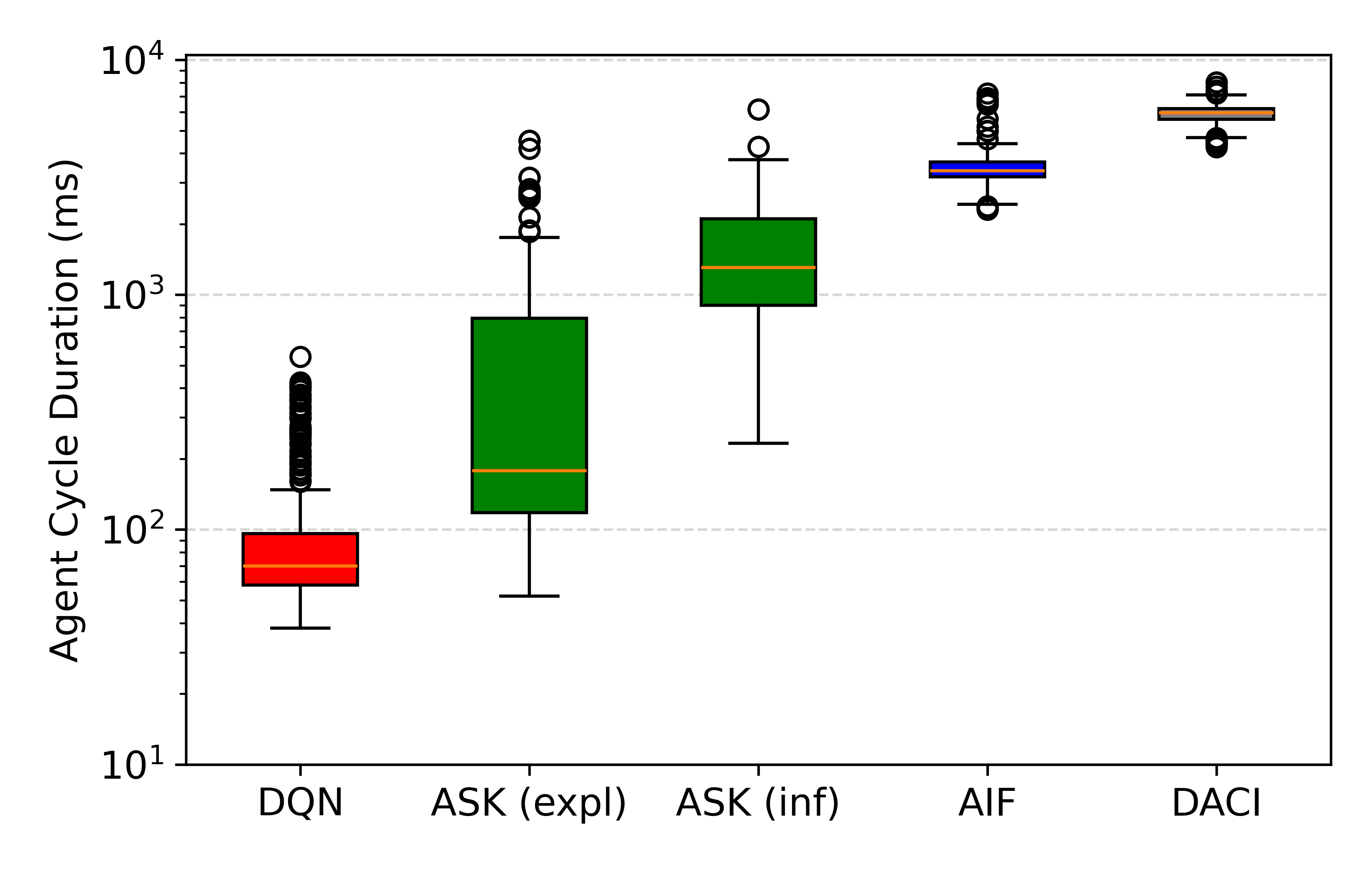}
        \caption{Duration of cycles ($log_{10}$) in milliseconds}
        \label{fig:loop_time}
    \end{subfigure}
    \caption{Scaling agents operating on the processing environment for 50 iterations (=250 seconds); 10 experiment repetitions per agent type to stabilize results}
    \label{fig:mainfig}
    \vspace{-10pt}
\end{figure}

Figure~\ref{fig:slo_fulfillment} shows the SLO fulfillment rate of all agents over the 50 steps of the experiment. The ASK agent requires approximately 20 iterations to train and stabilize, reaching an average SLO fulfillment rate of \textbf{0.868} over the final 10 steps. In contrast, the DQN agent, having been pre-trained in the simulation environment, maintains stable performance throughout the experiment and achieves an average of \textbf{0.753} in the last 10 iterations. 
Regarding the active inference-based methods, the AIF agent takes around 10 steps to stabilize its performance, ultimately reaching an average fulfillment rate of \textbf{0.704} during the final 10 steps.
Lastly, although the DACI agent exhibits steady improvement, it only achieves an average SLO fulfillment rate of \textbf{0.724} by the end of the experiment.


In terms of computational speed (Figure~\ref{fig:loop_time}) there is a considerable difference between the agents – note that the y-axis is plotted in log10 scale.
The DQN agent remains the fastest with a mean of \textbf{60} milliseconds, benefiting
from mature neural network optimizations. Our optimized AIF agent achieves
an average time of \textbf{1.7} seconds, demonstrating substantial improvement through vectorized inference over the original implementation with roughly \textbf{20} seconds. The DACI and ASK agents exhibit execution times of \textbf{2.8} seconds and \textbf{1.2} seconds respectively, highlighting the computational trade-offs inherent in their respective approaches. When merely exploring (expl) the ASK agent also took less time than during the actual numerical inference (inf). Notably, all agents (incl. the AIF) now operate within practical time constraints for the 5-second scaling interval, enabling real-time performance comparison across all four methods.


\section{Conclusion \& Future Work}\label{sec:conclusion}

This article highlights a key need for orchestration under resource limitations: multi-dimensional elasticity. By optimizing resource allocation and finding quality trade-offs, it supports flexible scaling behavior that improve Service Level Objectives (SLOs). To show the capabilities of different learning methods for solving such problems, we developed a real-world processing environment where two services compete for resources. We compared four different agents: an Active Inference (AIF) agent using \texttt{pymdp}, a Deep Q-Network (DQN), a Deep Active Inference (DACI) agent, and an agent using Analysis of Structural Knowledge (ASK).
Our evaluation shows that ASK achieved the highest SLO fulfillment (0.87), with AIF, DQN and DACI roughly taking the second spot together; AIF and ASK, which learn online, took about 10 and 20 rounds to converge respectively.
In terms of execution time, agents must infer a scaling policy in less than 5 seconds -- our orchestration interval. Our optimized AIF agent achieved an average of 1.7 seconds per iteration, showcasing our improvement of the pymdp library. However, it was the DQN agent that excelled with an average execution under 60 ms. While DACI exhibited the longest execution time, it is arguably the most promising approach as it combines theoretical foundations with practical scalability, which are a necessity in large-scale distributed systems.

We emphasize that we only present early-stage results. Besides further improvements to the neural architecture and the objective, careful theoretical analysis, extensive experimentation, and optimizations that can exploit parallel processing are works in progress. Moreover, future work will refine agents for distributed settings, explore service-hardware interaction, and develop tools for easier configuration and orchestration. We also plan to extend the framework to support a broader set of real-world use cases.

\section*{Acknowledgment}
This work is partially supported by CNS2023-144359 and the European Union NextGenerationEU/PRTR under MICIU/AEI/10.13039/501100011033, and partially by the European Union under (TEADAL, 101070186).
\vspace{-5pt}

\bibliographystyle{splncs04}
\bibliography{boris, references}

\end{document}